\documentclass[a4paper,conference]{IEEEtran}
\IEEEoverridecommandlockouts
\usepackage{multirow}
\usepackage{booktabs}
\usepackage{algorithm}
\usepackage{algorithmic}
\usepackage{cite}
\usepackage{amsmath,amssymb,amsfonts}
\usepackage{graphicx}
\usepackage{textcomp}
\usepackage{xcolor}
\usepackage{authblk}
\usepackage{algorithm}
\usepackage{algorithmic}
\usepackage{amsmath}
\usepackage{newfloat}
\usepackage{listings}
\def\BibTeX{{\rm B\kern-.05em{\sc i\kern-.025em b}\kern-.08em
    T\kern-.1667em\lower.7ex\hbox{E}\kern-.125emX}}
\begin{document}

\title{Mechanistic Understandings of Representation Vulnerabilities and Engineering Robust Vision Transformers\\
% {\footnotesize \textsuperscript{*}Note: Sub-titles are not captured for https://ieeexplore.ieee.org  and
}
% \author{\IEEEauthorblockN{Anonymous Authors}
% \IEEEauthorblockA{\textit{Anonymous Institution(s)} \\}
% }
\author[*]{Chashi Mahiul Islam
% \thanks{Corresponding Author: ci20l@fsu.edu}
}
\author[*]{Samuel Jacob Chacko}
\author[+]{Mao Nishino}
\author[*]{Xiuwen Liu}

\affil[*]{\textit{Department of Computer Science, Florida State University, Tallahassee, USA}}
\affil[+]{\textit{Department of Mathematics, Florida State University, Tallahassee, USA}}

\affil[ ]{\{ci20l@fsu.edu, sj21j@fsu.edu, mn21h@fsu.edu, liux@cs.fsu.edu\}}
\affil[ ]{Corresponding Author: ci20l@fsu.edu}

% \author{\IEEEauthorblockN{Anonymous Authors}
% \IEEEauthorblockA{\textit{Anonymous Institution(s)} \\}
% }
\maketitle

\begin{abstract}
While transformer-based models dominate NLP and vision applications, their underlying mechanisms to map the input space to the label space semantically are not well understood. In this paper, we study the sources of known representation vulnerabilities of vision transformers (ViT), where perceptually identical images can have very different representations and semantically unrelated images can have the same representation. Our analysis indicates that imperceptible changes to the input can result in significant representation changes, particularly in later layers, suggesting potential instabilities in the performance of ViTs. Our comprehensive study reveals that adversarial effects, while subtle in early layers, propagate and amplify through the network, becoming most pronounced in middle to late layers. This insight motivates the development of NeuroShield-ViT, a novel defense mechanism that strategically neutralizes vulnerable neurons in earlier layers to prevent the cascade of adversarial effects. We demonstrate NeuroShield-ViT's effectiveness across various attacks, particularly excelling against strong iterative attacks, and showcase its remarkable zero-shot generalization capabilities. Without fine-tuning, our method achieves a competitive accuracy of 77.8\% on adversarial examples, surpassing conventional robustness methods. Our results shed new light on how adversarial effects propagate through ViT layers, while providing a promising approach to enhance the robustness of vision transformers against adversarial attacks. Additionally, they provide a promising approach to enhance the robustness of vision transformers against adversarial attacks.

% The link to our code and data on GitHub will be included here in the published version.
\end{abstract}

\begin{IEEEkeywords}
representation vulnerability, adversarial attack, vision transformer, robust embedding
\end{IEEEkeywords}
\section{Introduction}

The advent of Vision Transformers (ViTs) \cite{dosovitskiy2020image} has revolutionized the field of computer vision, building upon the success of Transformer architectures in natural language processing \cite{vaswani2017attention}. Unlike traditional Convolutional Neural Networks (CNNs) \cite{krizhevsky2012imagenet}, ViTs leverage self-attention mechanisms applied to sequences of image patches, achieving state-of-the-art performance across various tasks such as image classification, semantic segmentation \cite{zheng2021rethinking}, and object detection \cite{carion2020end}. The versatility of ViTs has facilitated their integration into powerful multimodal models \cite{radford2021learning}, enabling advanced capabilities in zero-shot learning and cross-modal retrieval \cite{xu2023multimodal}. Despite their impressive performance, however, ViTs have shown vulnerabilities to adversarial attacks \cite{shao2021adversarial}, highlighting the need for robust defense mechanisms such as adversarial training \cite{madry2017towards,wang2019improving}. Recent efforts to reverse-engineer these complex networks \cite{dar2022analyzing, geva2022transformer, raghu2021vision} have provided valuable insights into their inner workings, paving the way for improved interpretability and control of their predictions. As ViTs continue to dominate the landscape of deep learning, understanding their mechanisms and addressing their vulnerabilities remains crucial for advancing the field of computer vision and developing more reliable AI systems.

To effectively address these vulnerabilities, it is essential to understand how representations are processed and propagated across the layers of Vision Transformers. Recent research has begun to explore the layer-wise behavior of ViTs, focusing on how information is encoded and transformed throughout the model \cite{liu2023understanding, vilas2024analyzing}. However, a significant need remains for mechanistic analysis, specifically in the context of adversarially perturbed images. Such analysis is vital for identifying the layers and blocks that contribute most to the model's vulnerabilities, providing insights that could lead to the development of more robust and resilient Vision Transformer architectures.

In this paper, we address this need by conducting a detailed, layer-wise analysis of Vision Transformers under adversarial perturbations. Our contributions are twofold:

\begin{itemize}
    \item \textbf{Mechanistic Analysis of Representation Vulnerabilities:} We systematically examine how adversarial perturbations impact the internal representations within Vision Transformers, focusing on the discrepancies in representations for perceptually identical images and the similarities in representations for semantically unrelated images.
\item \textbf{NeuroShied-ViT:} 
We introduce NeuroShield-ViT, a novel defense mechanism that mitigates representation vulnerabilities in Vision Transformers through dynamic activation modification by selectively neutralizing adversarial neurons. This method identifies and attenuates critical activations associated with adversarial (perturbed) inputs during inference without requiring adversarial training. To the best of our knowledge, we are the first to implement this approach in ViT for adversarial robustness.
\end{itemize}

\section{Related Work}
Transformers \cite{vaswani2017attention} have significantly advanced performance across multiple domains, including natural language processing \cite{devlin2018bert}, computer vision \cite{dosovitskiy2020image}, and speech recognition \cite{dong2018speech}, by efficiently capturing long-range dependencies through their self-attention mechanism and transformer architecture. Much of the ongoing research aims to refine Transformer models' internal representations \cite{devlin2018bert, biswas2022geometric, zou2023representation, raghu2021vision}, focusing on improving accuracy, robustness, and behavior control \cite{zhang2024towards}. These efforts have led to substantial improvements in model interpretability and performance across various tasks. The extension of Transformer models to visual tasks has led to the development of Vision Transformers (ViTs) \cite{dosovitskiy2020image}, which have been analyzed for their ability to capture class-specific features \cite{vilas2024analyzing} and their handling of input images within high-dimensional embedding spaces \cite{salman2024intriguing}. Our work builds on this research to mechanistically explore how adversarial attacks influence the layer-wise representations in ViTs.

Despite these advancements, ViTs remain vulnerable to adversarial attacks, presenting significant challenges for their deployment in security-critical applications. Several attack methods, effective across various deep neural network models, have been adapted to target ViTs, including optimization-based methods \cite{carlini2017towards, carlini2017adversarialexampleseasilydetected, moosavi2016deepfool}, gradient-based techniques \cite{goodfellow2014explaining, madry2017towards, kurakin2018adversarial, papernot2016limitations}, and transfer attacks \cite{papernot2016transferability}. Recent studies \cite{shao2021adversarial, kim2024exploring, wang2022understanding} have applied these techniques to investigate the adversarial robustness of ViTs, revealing unique vulnerabilities in their attention mechanisms and token-based processing. Notably, \cite{islam2024malicious} demonstrated how vulnerabilities in vision-language models' embeddings can enable manipulation of navigation paths, while \cite{wei2022towards} leveraged ViTs' self-attention mechanisms for transferable attacks. Other works, such as \cite{wang2024dual}, have explored dual-stage attacks for cross-architecture transferability, and \cite{baras2023quantattack} examined adversarial techniques specific to model quantization. Additionally, representation space-based techniques \cite{salman2024intriguing} have been proposed, including self-supervised learning for unlabeled data \cite{rando2022exploring} and enhanced attack transferability using token gradient regularization \cite{zhang2023transferable}. Novel optimization techniques, such as transferable triggers \cite{wallace2019universal} and gradient-regularized relaxation \cite{chacko2024adversarial}, can enhance ViT attack methodologies, bridging insights across modalities.

Alongside adversarial attack research, significant focus has been on developing defenses to enhance model robustness without sacrificing accuracy \cite{li2023trade}, as maintaining both robustness and performance remains a key challenge. Adversarial training, introduced by \cite{goodfellow2014explaining}, has been widely adopted in ViTs \cite{mo2022adversarial, fu2022patch}, with patch-based techniques \cite{liu2023understanding} and edge information utilization \cite{li2024harnessing} showing promising results in improving model resilience. Other defenses include random entangled tokens \cite{gong2024random}, prediction averaging over noisy inputs \cite{cohen2019certified}, and methods that reduce perturbation impact using small receptive fields and masking \cite{xiang2021patchguard}. Some strategies modify the representation space, such as robust representation sensors \cite{zhang2024towards} or activation guidance \cite{turner2023activation, liu2023context}, offering novel approaches to enhance model robustness. \cite{wei2024assessing} assessed model vulnerability using parameter pruning and low-rank weight approximations, revealing fundamental insights into model brittleness and the trade-offs between model complexity and robustness. Our method differs by introducing neuron neutralization in ViTs to enhance adversarial robustness, without requiring adversarial training and outperforming traditional methods with a fraction of the dataset, providing a more efficient and effective approach to adversarial defense.

\section{Analysis of Adversarial Effect Propagation in Vision Transformers}
In our objective of a mechanistic understanding of representation vulnerabilities in ViTs, we comprehensively analyze adversarial effect propagation through the model's architecture. Vision Transformers, which have demonstrated remarkable performance in various computer vision tasks, operate by dividing an image $\mathbf{x} \in \mathbb{R}^{H \times W \times C}$ into fixed-size patches $\mathbf{x}_p \in \mathbb{R}^{N \times (P^2 \cdot C)}$, linearly embedding these patches and processing them through a series of self-attention and feed-forward layers. Here, $(H, W)$ is the image resolution, $C$ is the number of channels, $(P, P)$ is the resolution of each patch, and $N = HW/P^2$ is the resulting number of patches. Despite their efficacy, ViTs, like other deep learning models, are susceptible to adversarial attacks. To probe these vulnerabilities mechanistically, we employ an iterative optimization-based adversarial attack that leverages a target image's representation embedding (CLS token) to create a perturbed version of a given input image. This approach allows us to precisely manipulate the model's internal representations, providing insight into the mechanisms of representation vulnerability. We analyze adversarial effect propagation through ViT architectures from three key perspectives: Global Token Representation Analysis, examining how adversarial perturbations affect the CLS token across layers; Spatial Token Perturbation Dynamics, investigating the localized effects on image patch embeddings; and Neuron-Specific Activation Shifts, delving into individual neuron responses to perturbed inputs. 

The ViT model can be described as a function $f_\theta: \mathbb{R}^{H \times W \times C} \rightarrow \mathbb{R}^K$, where $K$ is the number of classes and $\theta$ represents the model parameters. We analyze the model's behavior on both input images $\mathbf{x}$ and their perturbed counterparts $\mathbf{x}_{adv}$ defined by $\mathbf{x}_{adv} = \mathbf{x} + \boldsymbol{\delta}$, and $\boldsymbol{\delta}$ is a small perturbation designed to mislead the model. The image patches $\mathbf{x}_p \in \mathbb{R}^{N \times (P^2 \cdot C)}$ are linearly projected to obtain token embeddings $\mathbf{z}_0 \in \mathbb{R}^{N \times D}$, where $D$ is the embedding dimension. A learnable classification token $\mathbf{z}_\text{cls} \in \mathbb{R}^{1 \times D}$ is prepended to this sequence, resulting in the input sequence $\hat{\mathbf{z}}_0 = [\mathbf{z}_\text{cls}; \mathbf{z}_0] \in \mathbb{R}^{(N+1) \times D}$. This sequence is then processed through $L$ layers of self-attention and feed-forward networks, with each layer $l$ producing an output $\hat{\mathbf{z}}_l \in \mathbb{R}^{(N+1) \times D}$.

\textbf{Model, Dataset and Attack Method: }
Our analysis uses a ViT-base model \cite{rw2019timm} (12 layers, patch size 16$\times$16, input shape 224$\times$224) pretrained on ImageNet-1K \cite{deng2009imagenet}.  We selected a subset of 100 classes from ImageNet-1K, choosing 5 images per class. We randomly picked a target class and image for each image, then applied the adversarial attack, which utilizes Iterative Gradient-based Optimization (IGO) \cite{salman2024intriguing}. This attack aims to match the input image's representation with the target image's by minimal perturbation, chosen for its high effectiveness in confusing models (compared to other attacks; see Table~\ref{tab:performance}). We created 500 sample pairs of input and perturbed images, forming the basis for our study.

\subsubsection{Global Token Representation Analysis}
\begin{figure}[t]
\centering
\includegraphics[width=.95\columnwidth]{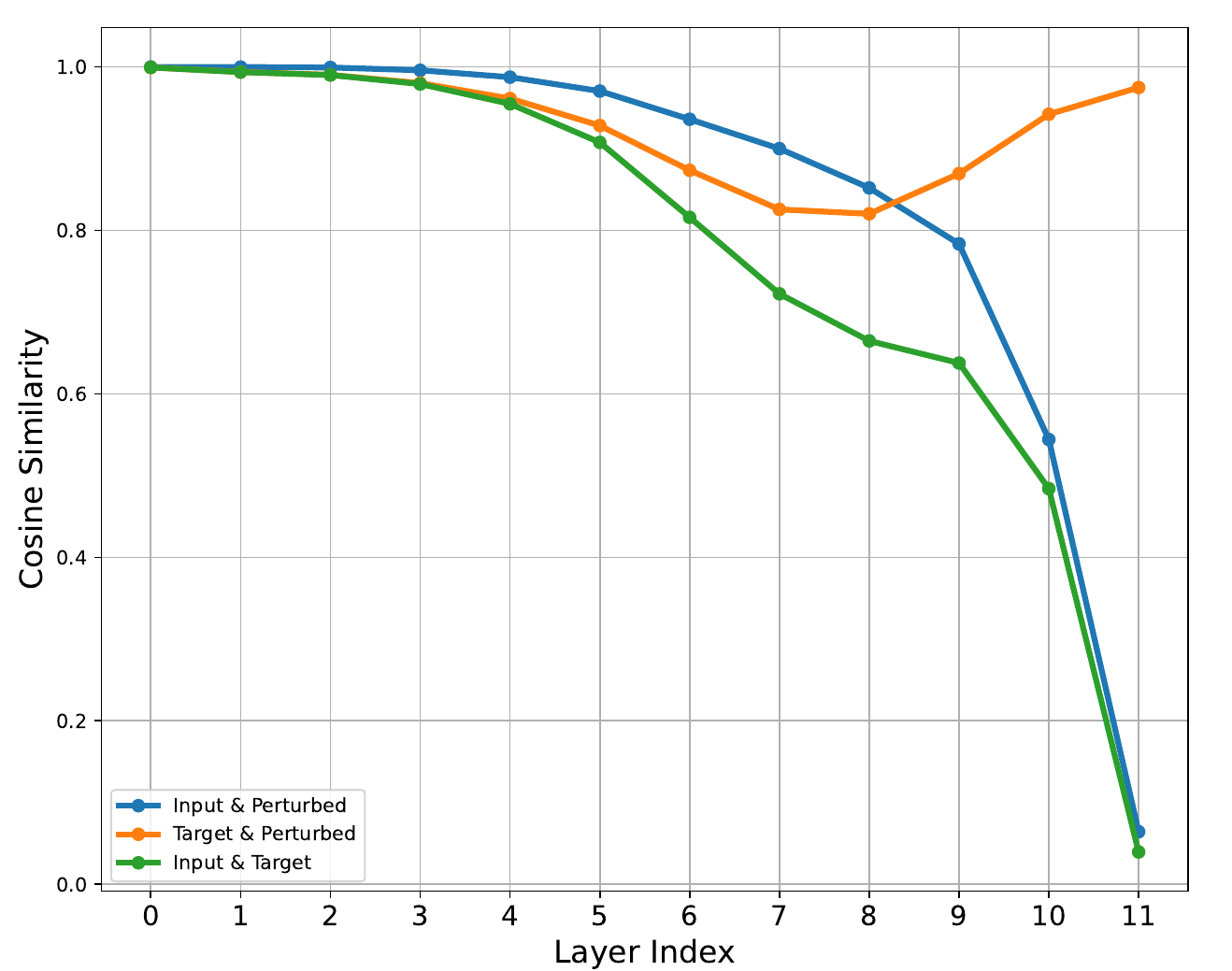}
\caption{Average layer-wise cosine similarity of CLS embeddings for input-perturbed, target-perturbed, and input-target image pairs across the ViT-B layers.}
\label{fig:global_token_representation}
\end{figure}
We examine the progression of the CLS token representation $\mathbf{z}_{cls}^l \in \mathbb{R}^D$ through the layers $l = 1, ..., L$ of the ViT, where $D$ is the embedding dimension. For each layer $l$, we compute the cosine similarity between the CLS token representations of clean (input) and adversarial (perturbed) images:
\begin{equation}
    \text{sim}(\mathbf{z}_{cls\_clean}^l, \mathbf{z}_{cls\_adv}^l) = \frac{\mathbf{z}_{cls\_clean}^l \cdot \mathbf{z}_{cls\_adv}^l}{\|\mathbf{z}_{cls\_clean}^l\| \|\mathbf{z}_{cls\_adv}^l\|}
\end{equation}

We analyze the similarity trajectory $S = \{s_1, ..., s_L\}$, where $s_l = \text{sim}(\mathbf{z}_{cls\_clean}^l, \mathbf{z}_{cls\_adv}^l)$, to identify critical layers where adversarial effects become most pronounced. Figure~\ref{fig:global_token_representation} illustrates CLS embeddings' average layer-wise cosine similarity for input-perturbed, target-perturbed, and input-target image pairs. Our analysis reveals a distinct pattern across the ViT layers:

\begin{itemize}
    \item Early layers (0-2): High similarity between input and perturbed images suggests that while adversarial perturbations are present, their effect on the model's representations is initially subtle.
    \item Middle layers (3-8): A gradual divergence becomes apparent, indicating an amplification of these perturbations.
    \item Later layers (9-11): A sharp decrease in similarity between input and perturbed representations is observed, coupled with an increase in similarity to the target class.
\end{itemize}

This pattern implies that the critical changes leading to misclassification are encoded early in the network in latent form, despite the apparent high similarity. These findings suggest that addressing adversarial effects in earlier layers could be more effective in preserving the model's overall performance, as it may prevent the propagation and amplification of subtle perturbations through the network.

\subsubsection{Spatial Token Perturbation Dynamics}
\begin{figure*}[t]
\centering
\includegraphics[width=1\textwidth]{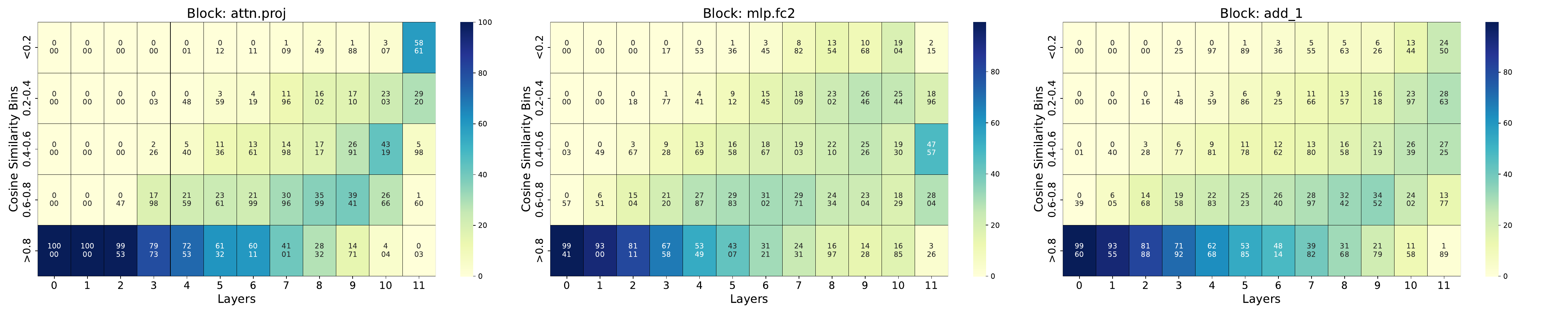}
\caption{Heatmaps illustrating the distribution of cosine similarities for image token embeddings in attention projection (attn.proj), MLP feed-forward (mlp.fc2), and residual connection (add\_1) blocks across ViT-B layers.}
\label{fig:spatial_token_perturbation}
\end{figure*}
Extending our analysis from global representations to local features, we investigate the embeddings of individual image patches $\mathbf{z}_i^l \in \mathbb{R}^D$, where $i = 1, ..., N$ and $N$ is the number of patches. For each transformer block, we focus on three key components:

\begin{enumerate}
    \item Attention Projection (att.proj): $\mathbf{Z}_{att}^l = \text{MultiHead}(\mathbf{Q}^l, \mathbf{K}^l, \mathbf{V}^l)\mathbf{W}^O$
    \item Feed-Forward Network Output (mlp.fc2): $\mathbf{Z}_{ffn}^l = \text{FFN}(\mathbf{Z}_{att}^l)$
    \item Residual Connection Output (add\_1): $\mathbf{Z}_{res}^l = \mathbf{Z}_{in}^l + \mathbf{Z}_{ffn}^l$
\end{enumerate}

For each component, we compute pairwise cosine similarities between clean and adversarial embeddings:
\begin{equation}
    \text{sim}(\mathbf{z}_{i\_clean}^l, \mathbf{z}_{i\_adv}^l) = \frac{\mathbf{z}_{i\_clean}^l \cdot \mathbf{z}_{i\_adv}^l}{\|\mathbf{z}_{i\_clean}^l\| \|\mathbf{z}_{i\_adv}^l\|}
\end{equation}

Figure~\ref{fig:spatial_token_perturbation} presents heatmaps illustrating the distribution of cosine similarities for image token embeddings in the three key blocks. These heatmaps reveal a nuanced progression of adversarial effects through the network, complementing our earlier CLS token analysis:

\begin{itemize}
    \item Early layers (0-2): High similarities across all blocks, with 80-100\% of embeddings in the 0.8-1.0 similarity range, confirm the subtle nature of initial perturbations observed in the CLS token analysis.
    \item Middle layers (3-8): A gradual divergence becomes apparent, particularly in MLP block, indicating the onset of perturbation amplification. This aligns with the divergence observed in the CLS token representations at these layers.
    \item Later layers (9-11): Significant divergence is observed, especially in attention projection, where a substantial portion of embeddings fall into lower similarity ranges.
\end{itemize}

Interestingly, residual connections (add\_1) maintain higher similarities even in later layers, suggesting their role in preserving some original input information. This block-specific analysis reveals that adversarial effects are not uniform across the network components, with MLP blocks appearing most susceptible to perturbations.

\subsubsection{Neuron Sensitivity and Activation Pattern Analysis}
\begin{figure}[t]
\centering
\includegraphics[width=1\columnwidth]{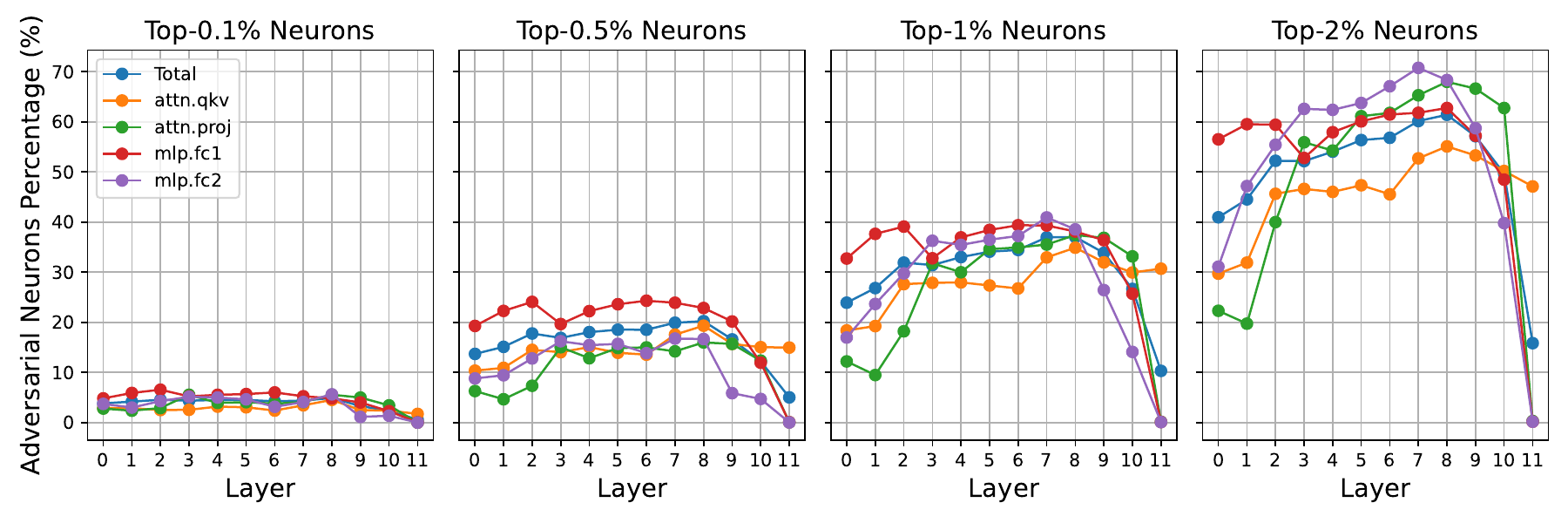} % Adjust the width as needed
\caption{Top-p\% adversarial neurons per representation responsible for adversarial effects across layers.}
\label{fig:adversarial_neurons}
\end{figure}
We conduct a granular analysis of individual neuron activations $\mathbf{a}_j^l \in \mathbb{R}$ across different layers $l$ and blocks of the ViT. For each neuron $j$ in layer $l$, we compute its importance $I_j^l$ using the product of activations and gradients:

\begin{equation}
    I_j^l = \sum_{i=1}^N (\mathbf{a}_{i,j}^l \cdot \mathbf{g}_{i,j}^l)
\end{equation}

For each image patch $\mathbf{x}_p \in \mathbb{R}^{P^2 \cdot C}$, we identify the top-$p\%$ neurons $\mathcal{N}_p^l$ based on $|I_{i,j}^l|$ for both input and perturbed inputs. The set of adversarial neurons $\mathcal{A}_p^l = \mathcal{N}_{p\_adv}^l \setminus \mathcal{N}_{p\_clean}^l$ represents neurons uniquely important for perturbed examples. Figure~\ref{fig:adversarial_neurons} illustrates the percentage of adversarial neurons across layers for $p \in \{0.1\%, 0.5\%, 1\%, 2\%\}$. We observe that adversarial effects remain localized in early layers (0-2), with a low percentage of adversarial neurons across all $p$ values. However, a significant spread begins in middle layers (3-8), particularly for $p \geq 1\%$, reaching 30-40\% adversarial neurons. This spread amplifies in later layers (9-11), peaking at 60-70\% for $p=2\%$, before a sharp decline in the final layer.

Notably, MLP blocks (mlp.fc1, mlp.fc2) show earlier and more pronounced spreading compared to attention blocks (attn.qkv, attn.proj). These findings align with our CLS token and spatial token analyses, confirming that adversarial perturbations, while subtle in early layers, propagate and amplify through the network. The localization of adversarial effects in early layers suggests that controlling the impact of these initially affected neurons could help inhibit the cascade effect observed in later layers. Intriguingly, we observe a marked drop in the percentage of unique adversarial neurons in the final 1-2 layers. This phenomenon can be attributed to a convergence of representations and the model's increased focus on class-specific features near the decision boundary \cite{vilas2024analyzing}. Additionally, architectural elements like normalization \cite{ba2016layer} and residual connections \cite{he2016deep} may play a role in stabilizing activations.
\section{Methodology}
\subsection{Defense Method: Selective Neuron Neutralization}

We propose NeuroShield-ViT (Algorithm~\ref{alg:selective_neuron_neutralization}), a novel defense method that mitigates adversarial perturbations by strategically modifying the activations of vulnerable neurons in Vision Transformers (ViTs). The algorithm begins by loading a pre-trained ViT and a dataset of input-perturbed image pairs. It then calculates neuron importance using both activations and gradients, identifying the top-$p\%$ most important neurons per representation for input and perturbed images. By computing the set difference between these neuron sets, NeuroShield-ViT isolates neurons uniquely significant to adversarial examples.

The core of NeuroShield-ViT lies in its selective neutralization approach. It applies a neutralization coefficient (within the range $[0, 0.5)$) to modify the activations of the identified adversarial neurons across specified layers and blocks. Finally, the algorithm performs a forward pass on the modified network to obtain an updated, more robust classification. This targeted approach effectively mitigates the impact of neurons contributing most significantly to representation vulnerabilities, thereby enhancing the overall robustness of ViT models against various attacks.

\begin{algorithm}[t!]
\caption{NeuroShield-ViT  - Enhancing Robustness through Selective Neuron Neutralization}
\label{alg:selective_neuron_neutralization}
\begin{algorithmic}[1]
\REQUIRE Pre-trained ViT model $M$, Dataset $D$ (input $I_{\text{input}}$ and perturbed image $I_{\text{adv}}$ pairs), Top-p percentage $p$, Neutralization Coefficient $\alpha$ $\in [0, 0.5)$, Layer list $L$, Block list $B$
\ENSURE Final classification after selective neuron neutralization
\FORALL{$(I_{\text{input}}, I_{\text{adv}}) \in D$}
    \STATE \textbf{Calculate Important Neurons:}
    \STATE Compute neuron importance for $I_{\text{input}}$ using activations and gradients:

    \begin{equation}
        I_j^l = \sum_{i=1}^N (\mathbf{a}_{i,j}^l \cdot \mathbf{g}_{i,j}^l)
    \end{equation}
    
    where $\mathbf{a}_{i,j}^l$ is the activation and $\mathbf{g}_{i,j}^l$ is the gradient for neuron $j$ in layer $l$ for patch $i$.

    \STATE \quad Repeat the process for $I_{\text{adv}}$.
    \STATE \quad Select top-$p\%$ neurons by importance for both $I_{\text{input}}$ and $I_{\text{adv}}$.
\ENDFOR
\STATE \textbf{Calculate Set Difference:}
\STATE \quad $\forall l \in L, b \in B$: Compute set difference of important neurons between $I_{\text{adv}}$ and $I_{\text{input}}$.
\STATE \quad Store neurons unique to $I_{\text{adv}}$ (contributing to representation vulnerability).
\FORALL{$l \in L$}
    \FORALL{$b \in B$}
        \STATE \textbf{Apply Neutralization:}
        \STATE \quad Modify activations of identified neurons with coefficient.
            \begin{equation}
        \hat{\mathbf{a}}_{i,j}^l = \alpha \cdot \mathbf{a}_{i,j}^l
    \end{equation}
    \ENDFOR
\ENDFOR
\STATE \textbf{Perform Forward Pass:} Obtain final classification $C_f$ post-neutralization.
\STATE \textbf{Output} $C_f$
\end{algorithmic}
\end{algorithm}

\section{Experiments and Results}

\subsection{Experimental Setup}

\textbf{Datasets and Models:}
We conducted our experiments on a variety of datasets, including mini ImageNet-1K \cite{deng2009imagenet} (a 100-class subset of ImageNet-1K), Imagenette \cite{imagenette}, and CIFAR-10 \cite{krizhevsky2010cifar}. Given the time-intensive process of generating adversarial samples for each experiment and attack scenario, we randomly selected 200 samples from each dataset to generate adversarially perturbed images for testing. The same number of input-perturbed image pairs were also generated for the purpose of determining neuron importance, which we call NeuroShield Calibration Set (NCS).

Our analysis encompasses four Vision Transformer variants: ViT-S, ViT-B, DeiT-S, and DeiT-B \cite{touvron2021training}. 
% All models employ a 16x16 patch size with 224x224 input dimensions. 
Unless specified otherwise, ViT-B serves as our primary model for detailed studies.

\textbf{Adversarial Attacks:}
We employ a range of adversarial attacks to evaluate model robustness:
\begin{itemize}
    \item Fast Gradient Sign Method (FGSM)
    \item Projected Gradient Descent (PGD-20 and PGD-100)
    \item Iterative Gradient Optimization (IGO)
\end{itemize}

For the IGO attack, we use a learning rate of 0.1 and terminate the process after 2000 iterations or when the cosine similarity of the final representation reaches 0.97, and the L2 distance between images is $\leq 36$.

\textbf{Implementation and Evaluation:}
Our framework is implemented in PyTorch, leveraging 2x NVIDIA RTX A5000 GPUs for faster computation. We use accuracy as our primary evaluation metric across different models, attack methods, and experimental studies.

\subsection{NeuroShield-ViT Performance Evaluation}
For these experiments, we utilized only 25\% of the NCS data, selecting  0.5 as the top\_p value and 0.1 as the neutralization coefficient. Table~\ref{tab:performance} presents the performance of NeuroShield-ViT across various Vision Transformer variants and datasets under different adversarial attacks. The results demonstrate that NeuroShield-ViT significantly enhances the robustness of Vision Transformers against a range of adversarial attacks, particularly for stronger attacks like IGO, PGD-20, and PGD-100. For instance, on the ImageNet1K (100) dataset, NeuroShield-ViT improves the accuracy of ViT-B from 0\% to 71.6\% against IGO attacks and from 11.7\% to 71.3\% against PGD-100 attacks. Similar improvements are observed across all model variants and both datasets. Notably, the method maintains a reasonable trade-off with natural accuracy, showing only modest decreases in most cases and even slight improvements for one configuration on the Imagenette dataset. We hypothesize that the modest decrease in natural accuracy could be due to the neutralization mechanism subtly altering the model's decision boundary. Interestingly, the performance against FGSM attacks shows minimal improvement or slight decreases in some cases. This could be attributed to the relatively simple nature of FGSM \cite{goodfellow2014explaining} attacks, which may not significantly alter the critical neurons targeted by NeuroShield-ViT, or potentially due to a gradient masking effect induced by our method. Despite this, the substantial improvements against more sophisticated attacks underscore the effectiveness of NeuroShield-ViT in enhancing the adversarial robustness of Vision Transformers, even when using a limited amount of NCS data. On CIFAR-10 with ViT-B/16-224, NeuroShield-ViT significantly improved robustness, increasing accuracy from 23\% to 50\% for PGD-100 and 21\% to 49\% for PGD-20 attacks.

\begin{table*}[ht]
\caption{Performance (\%) of different methods on ViT variants across various datasets under different adversarial attacks. The best results are in bold.}
\label{tab:performance}
\centering

\setlength{\tabcolsep}{1.8mm} % Adjust column separation
\small % Set font size to 10-point
% \resizebox{\textwidth}{!}{%
\begin{tabular}{lccccccccccc}
\toprule
\textbf{Model} & \textbf{Method} & \multicolumn{5}{c}{\textbf{ImageNet1K (100)}} & \multicolumn{5}{c}{\textbf{Imagenette}} \\
\cmidrule(lr){3-7} \cmidrule(lr){8-12}
& & \textbf{Natural} & \textbf{FGSM} & \textbf{IGO} & \textbf{PGD-20} & \textbf{PGD-100} & \textbf{Natural} & \textbf{FGSM} & \textbf{IGO} & \textbf{PGD-20} & \textbf{PGD-100} \\
\midrule
\multirow{2}{*}{DeiT - S} 
& W/O Defense & \textbf{84.7} & 62.0 & 0.0 & 10.1 & 6.2 & \textbf{81.0} & 61.4 & 0.0 & 11.1 & 5.2 \\
& \textbf{NeuroShield-ViT} & 78.1 & 62.0 & \textbf{62.7} & \textbf{60.3} & \textbf{60.4} & 76.5 & \textbf{62.8} & \textbf{68.9} & \textbf{60.1} & \textbf{63.5} \\
\midrule
\multirow{2}{*}{DeiT - B} 
& W/O Defense & \textbf{79.0} & 68.0 & 2.4 & 24.1 & 10.0 & \textbf{78.66} & 55.3 & 0.0 & 17.7 & 6.8 \\
& \textbf{NeuroShield-ViT} & 72.7 & 68.0 & \textbf{64.3} & \textbf{44.2} & \textbf{52.3} & 75.66 & \textbf{62.3} & \textbf{65.7} & \textbf{50.0} & \textbf{59.3} \\
\midrule
\multirow{2}{*}{ViT - S} 
& W/O Defense & \textbf{81.5} & \textbf{54.2} & 0.0 & 11.3 & 1.0 & 79.37 & 57.5 & 0.0 & 12.5 & 7.5 \\
& \textbf{NeuroShield-ViT} & 71.0 & 53.2 & \textbf{24.0} & \textbf{40.7} & \textbf{33.0} & \textbf{80.6} & \textbf{62.5} & \textbf{25.0} & \textbf{50.1} & \textbf{42.5} \\
\midrule
\multirow{2}{*}{ViT - B} 
& W/O Defense & \textbf{88.0} & 68.0 & 0.0 & 20.2 & 11.7 & \textbf{85.2} & 53.6 & 0.0 & 23.4 & 15.6 \\
& \textbf{NeuroShield-ViT} & 85.7 & \textbf{72.3} & \textbf{71.6} & \textbf{70.3} & \textbf{71.3} & 84.5 & \textbf{60.0} & \textbf{77.8} & \textbf{62.2} & \textbf{72.3} \\
\bottomrule
\end{tabular}%
% }
\end{table*}

\subsection{Ablation Studies}

\textbf{Layer-wise Neutralization Analysis:}
\begin{figure}[t]
\centering
\includegraphics[width=1\columnwidth]{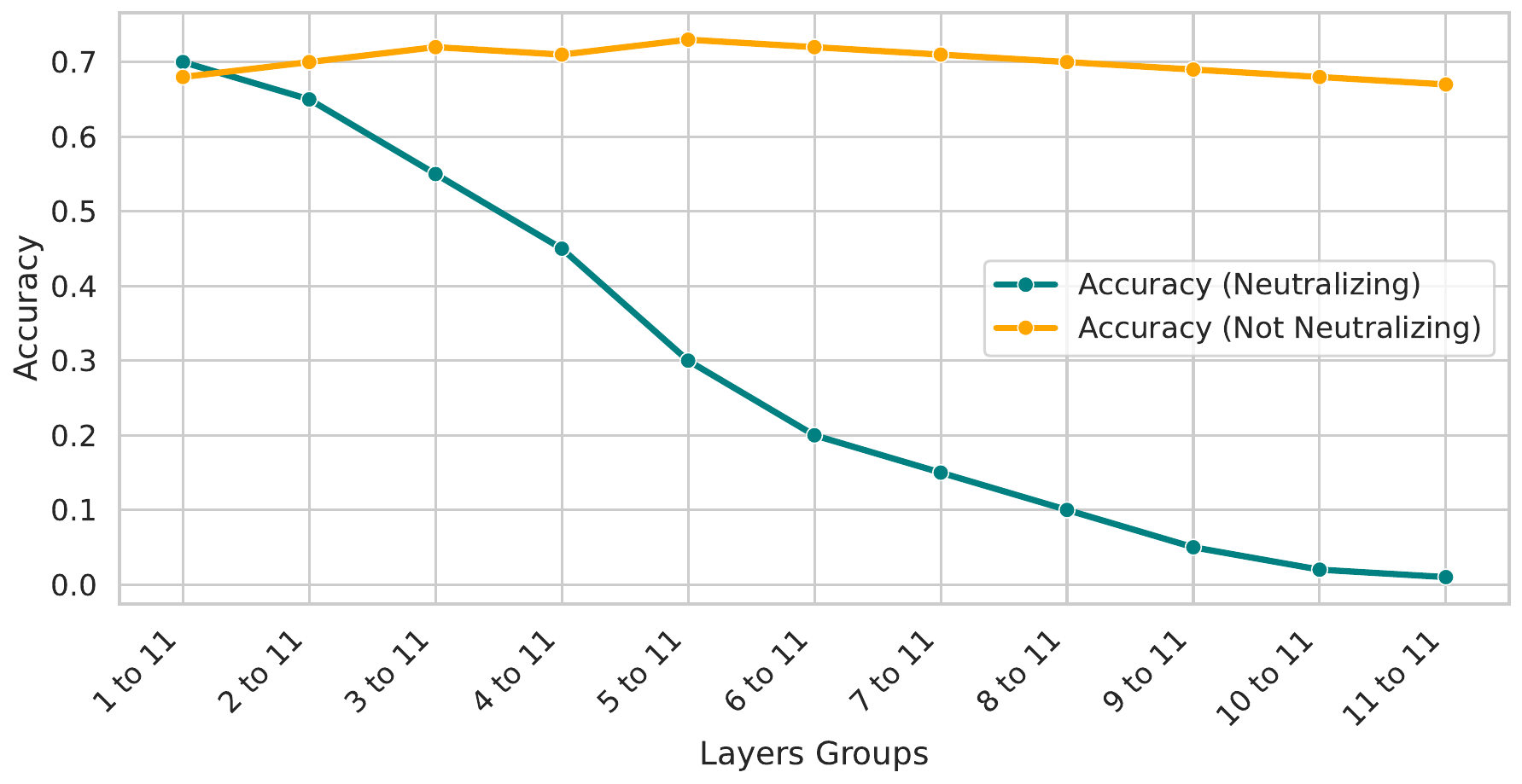}
\caption{Impact of layer-wise neutralization on model accuracy}
\label{fig:layerwise_inclusion_exlusion}
\end{figure}

Figure~\ref{fig:layerwise_inclusion_exlusion} illustrates the impact of layer-wise neutralization on model accuracy. The results demonstrate two key trends: (1) When neutralizing layers from the end, accuracy remains near zero until earlier layers are included, with significant improvements starting from layers 3-11. This suggests that neutralizing only later layers is ineffective. (2) Conversely, neutralizing only the first layer (0) already yields high accuracy, which remains stable as more layers are neutralized. This pattern aligns with our previous analyses of global representation, image token, and neuron sensitivity, confirming that addressing adversarial effects in earlier layers effectively prevents their propagation through the network. The findings strongly suggest that focusing neutralization efforts on early layers is most effective for maintaining model performance while mitigating adversarial effects, as these layers play a crucial role in stopping the spread of adversarial perturbations.

\subsection{Sensitivity Analysis}
\textbf{Data Percentage Impact:}

\begin{figure}[t]
\centering
\includegraphics[width=1\columnwidth]{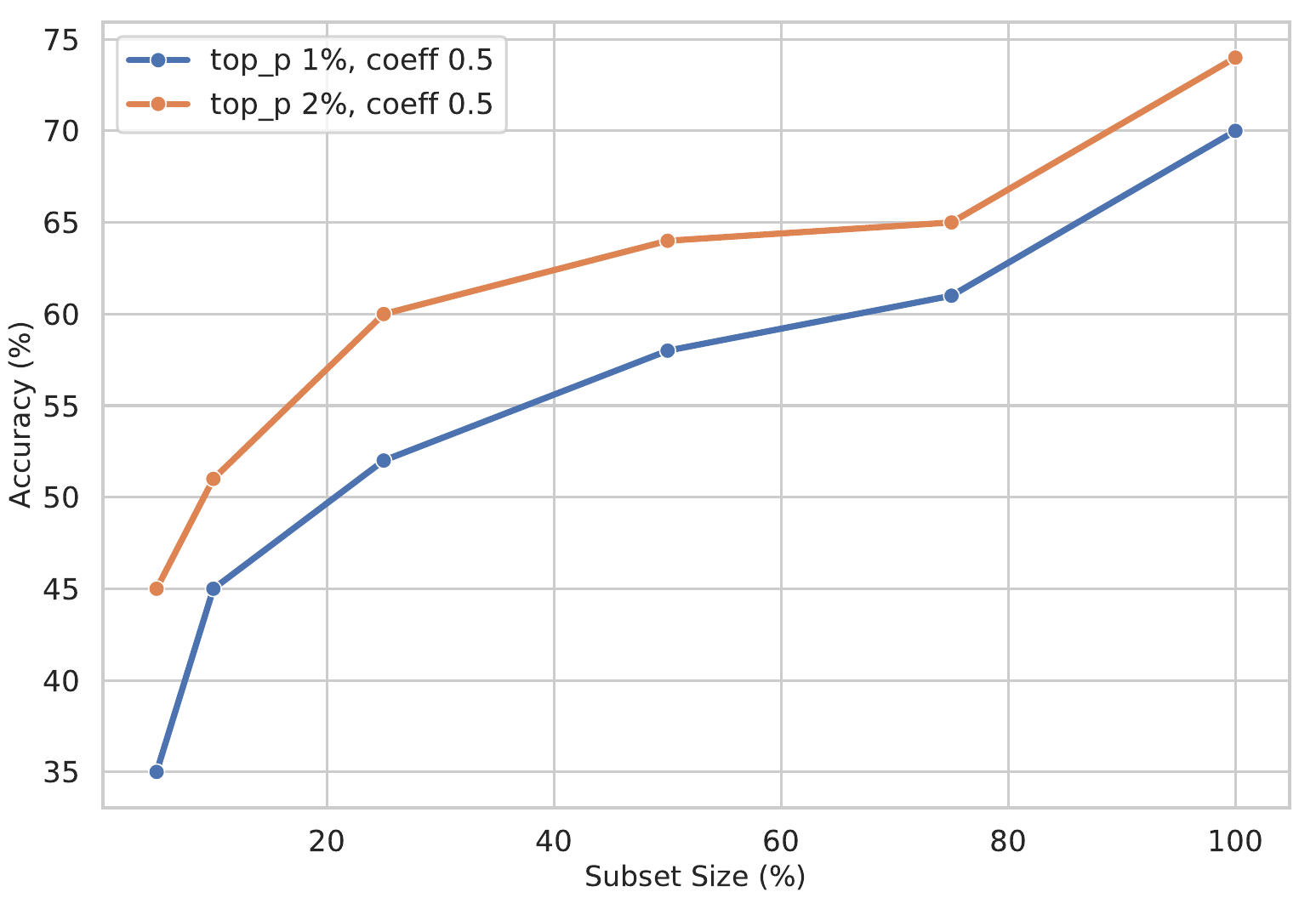}
\caption{Impact of the subset size (\%) of NCS on the model accuracy.}
\label{fig:susbet_size_vs_accuracy}
\end{figure}

Figure~\ref{fig:susbet_size_vs_accuracy} demonstrates the effect of increasing subset size on NeuroShield-ViT's performance. Accuracy consistently improves as the subset size grows from 5\% to 100\%, with top-p 2\% outperforming top-p 1\% across all sizes. The improvement is more pronounced for smaller subsets (5-20) and gradually levels off for larger ones (60-100). This trend shows a trade-off between data amount and effectiveness: initially, small increases in subset size lead to significant gains, but as the subset nears 100\%, the benefits become smaller. These findings suggest that NeuroShield-ViT can achieve robust performance without requiring the full dataset, offering a balance between computational efficiency and adversarial robustness.

\textbf{Parameter Sensitivity Evaluation:}  
We evaluated the parameter sensitivity of NeuroShield-ViT by analyzing the impact of subset size, top-p values, and neutralization coefficients on ViT-B/16-224 accuracy for ImageNet1k. Smaller subsets (e.g., size 5) achieve higher accuracy with lower top-p values (0.5\%-1.0\%) and moderate neutralization coefficients (0.1-0.2). Larger subsets and the full dataset perform better with higher top-p values, with optimal performance observed at 2.0\% top-p and a neutralization coefficient of 0.5. However, for larger subsets, using high top-p values with low neutralization coefficients reduces accuracy, likely due to over-neutralization of important features. These findings underscore the importance of carefully tuning parameters to achieve a balance between adversarial robustness and model performance.

\subsection{Generalization Study}
\textbf{Cross-class Generalization Assessment:}
Table~\ref{tab:imagenet1k_comparison} demonstrates NeuroShield-ViT's remarkable zero-shot generalization capabilities. Despite being neutralized on the 10-class Imagenette NCS data, NeuroShield-ViT shows significant improvements in robustness when applied to a distinct 100-class subset of ImageNet1K. Notably, for the Iterative attack, accuracy increases from 0\% to 71\%, and for PGD-100, from 11.7\% to 60\%. These results underscore NeuroShield-ViT's ability to identify and neutralize adversarial neurons that generalize across diverse class sets, suggesting that the method captures fundamental aspects of representation vulnerabilities in Vision Transformers. This cross-dataset efficacy highlights NeuroShield-ViT's potential as a versatile and powerful defense mechanism, capable of providing robust protection even for previously unseen classes.

\begin{table}[t]
\caption{Imagenet1k: ZeroShot performance (\%) with and without (W/O) NeuroShield-ViT across different attack methods. Neuron neutralization done with Imagenette subset.}
\label{tab:imagenet1k_comparison}
\centering
\setlength{\tabcolsep}{1.7mm} % Adjust column separation
\small % Set font size to 10-point
\begin{tabular}{lcccc}
\toprule
\textbf{Defense} & \textbf{FGSM} & \textbf{Iterative} & \textbf{PGD-20} & \textbf{PGD-100} \\
\midrule
W/O Defense & 68.0  & 0.0  & 20.2  & 11.7  \\
\textbf{NeuroShield-ViT} & 67.7  & \textbf{71.0}  & \textbf{62.2}  & \textbf{60.0}  \\
\bottomrule
\end{tabular}
\end{table}

\subsection{Comparative Analysis}
\begin{table}[ht]
\caption{Performance (\%) comparison with Adversarial Training (AT) and Randomized Smoothing (RS) on Different Datasets under Adversarial Attacks}
\label{tab:comparison}
\centering
\setlength{\tabcolsep}{2mm} % Adjust column separation
\small % Set font size to 10-point
\begin{tabular}{lcccc}
\toprule
\textbf{Attack} & \textbf{Method} & \textbf{ImageNet1K} & \textbf{Imagenette} \\
\midrule
\multirow{3}{*}{IGO} & W/O Defense & 0.0 & 0.0 \\
& AT & 5.0 & 8.0  \\
& RS & 45.8 & 46.0 \\
& NeuroShield-ViT & \textbf{71.6} & \textbf{77.8} \\
\midrule
\multirow{3}{*}{PGD-100} & W/O Defense & 20.2 & 15.6 \\
& AT & 27.5 & 29.0 \\
& RS & 45.9 & 40.4 \\
& NeuroShield-ViT & \textbf{70.3} & \textbf{72.3} \\
\bottomrule
\end{tabular}
\end{table}
Table~\ref{tab:comparison} presents a comparative analysis of NeuroShield-ViT against conventional adversarial training (AT) and randomized smoothing. Adversarial training \cite{goodfellow2014explaining}, widely considered a standard defense method \cite{costa2024deep}, shows minimal improvement against IGO and PGD-100 attacks on both ImageNet1K and Imagenette datasets, achieving only 5.0\% and 8.0\% accuracy on IGO attacks and 27.5\% and 29.0\% on PGD-100 attacks respectively, despite using the full NCS set for 20 epochs of training with a learning rate of 0.001. Randomized smoothing \cite{cohen2019certified}, which creates a provably robust classifier by adding Gaussian noise during inference, provides better robustness than AT with accuracy improvements of 45.8\% and 46.0\% on IGO attacks and 45.9\% and 40.4\% on PGD-100 attacks for ImageNet1K and Imagenette respectively. In contrast, NeuroShield-ViT, utilizing only 25\% of the NCS set, achieves significantly higher accuracy, with improvements of up to 71.6\% and 77.8\% against IGO attacks and 70.3\% and 72.3\% against PGD-100 attacks. This substantial performance gain, coupled with reduced computational requirements, underscores NeuroShield-ViT's efficiency. While recent improvements like ARD and PRM \cite{mo2022adversarial} have shown incremental average gains of 0.81\% and 1.15\%, respectively, over standard AT, NeuroShield-ViT's performance far surpasses these advancements and other certification methods like randomized smoothing, demonstrating its potential as a robust and efficient defense mechanism against adversarial attacks on Vision Transformers.

\section{Discussion}
Our comprehensive analysis of representation vulnerabilities in Vision Transformers reveals a critical pattern: adversarial effects, while subtle in early layers, propagate and amplify through the network, becoming most pronounced in middle to late layers. This insight underpins the success of NeuroShield-ViT, which strategically neutralizes vulnerable neurons in earlier layers to prevent the cascade of adversarial effects. Compared to state-of-the-art methods like adversarial training, NeuroShield-ViT demonstrates superior robustness across various attacks, particularly excelling against strong iterative attacks. Its effectiveness in zero-shot scenarios further highlights its capacity to capture fundamental aspects of representation vulnerabilities. However, the method's reliance on pre-identified vulnerable neurons may limit its adaptability to novel attack types. Future work should explore dynamic neuron identification techniques and investigate the method's applicability to other transformer architectures beyond vision tasks. Additionally, analyzing and neutralizing specific blocks within layers, rather than whole-layer neutralization, could potentially reduce computational costs and increase the method's speed. 

\section{Conclusion}
This study provides a comprehensive mechanistic understanding of representation vulnerabilities in Vision Transformers (ViTs) and introduces NeuroShield-ViT, an effective defense mechanism against adversarial attacks.
 By focusing on early-layer resilience and selective neuron manipulation, our approach demonstrates the potential for creating more secure AI systems without extensive retraining. The principles underlying NeuroShield-ViT extend beyond computer vision, offering insights for improving model reliability across various domains of machine learning. As AI continues to permeate critical applications, these findings pave the way for developing more trustworthy and dependable systems, potentially influencing the design of future neural network architectures to prioritize inherent robustness against adversarial threats.

\onecolumn
\appendix
% \twocolumn
\section{Appendix 1}
\subsection{Parameter Evaluation}
\begin{table*}[ht]
\centering
\setlength{\tabcolsep}{4.5mm} % Adjust column separation
\small % Set font size to 10-point
% \resizebox{\textwidth}{!}{%
\begin{tabular}{cc}
\toprule
\multicolumn{2}{c}{\textbf{Imagenet1k Accuracy (\%)}} \\
\midrule
\begin{tabular}{lcccc}
\toprule
\multirow{2}{*}{\textbf{Top-p}} & \multicolumn{4}{c}{\textbf{Subset 5}} \\
\cmidrule(lr){2-5}
& \textbf{0} & \textbf{0.1} & \textbf{0.2} & \textbf{0.5} \\
\midrule
\textbf{0.5\%} & 65.0 & 54.0 & 48.0 & 16.0 \\
\textbf{1.0\%} & 64.0 & 64.0 & 61.0 & 35.0 \\
\textbf{2.0\%} & 67.0 & 68.0 & 67.0 & 45.0 \\
\bottomrule
\end{tabular} &
\begin{tabular}{lcccc}
\toprule
\multirow{2}{*}{\textbf{Top-p}} & \multicolumn{4}{c}{\textbf{Subset 10}} \\
\cmidrule(lr){2-5}
& \textbf{0} & \textbf{0.1} & \textbf{0.2} & \textbf{0.5} \\
\midrule
\textbf{0.5\%} & 66.0 & 64.0 & 59.0 & 42.0 \\
\textbf{1.0\%} & 66.0 & 64.0 & 66.0 & 45.0 \\
\textbf{2.0\%} & 61.0 & 66.0 & 67.0 & 51.0 \\
\bottomrule
\end{tabular} \\
\midrule
\begin{tabular}{lcccc}
\toprule
\multirow{2}{*}{\textbf{Top-p}} & \multicolumn{4}{c}{\textbf{Subset 25}} \\
\cmidrule(lr){2-5}
& \textbf{0} & \textbf{0.1} & \textbf{0.2} & \textbf{0.5} \\
\midrule
\textbf{0.5\%} & 67.0 & 68.0 & 65.0 & 53.0 \\
\textbf{1.0\%} & 61.0 & 62.0 & 65.0 & 52.0 \\
\textbf{2.0\%} & 62.0 & 65.0 & 65.0 & 60.0 \\
\bottomrule
\end{tabular} &
\begin{tabular}{lcccc}
\toprule
\multirow{2}{*}{\textbf{Top-p}} & \multicolumn{4}{c}{\textbf{Subset 50}} \\
\cmidrule(lr){2-5}
& \textbf{0} & \textbf{0.1} & \textbf{0.2} & \textbf{0.5} \\
\midrule
\textbf{0.5\%} & 65.0 & 68.0 & 67.0 & 58.0 \\
\textbf{1.0\%} & 62.0 & 62.0 & 64.0 & 58.0 \\
\textbf{2.0\%} & 53.0 & 55.0 & 61.0 & 64.0 \\
\bottomrule
\end{tabular} \\
\midrule
\begin{tabular}{lcccc}
\toprule
\multirow{2}{*}{\textbf{Top-p}} & \multicolumn{4}{c}{\textbf{Subset 75}} \\
\cmidrule(lr){2-5}
& \textbf{0} & \textbf{0.1} & \textbf{0.2} & \textbf{0.5} \\
\midrule
\textbf{0.5\%} & 64.0 & 65.0 & 69.0 & 61.0 \\
\textbf{1.0\%} & 57.0 & 63.0 & 67.0 & 61.0 \\
\textbf{2.0\%} & 43.0 & 53.0 & 58.0 & 65.0 \\
\bottomrule
\end{tabular} &
\begin{tabular}{lcccc}
\toprule
\multirow{2}{*}{\textbf{Top-p}} & \multicolumn{4}{c}{\textbf{Subset 100}} \\
\cmidrule(lr){2-5}
& \textbf{0} & \textbf{0.1} & \textbf{0.2} & \textbf{0.5} \\
\midrule
\textbf{0.5\%} & 61.0 & 65.0 & 67.0 & 67.0 \\
\textbf{1.0\%} & 54.0 & 59.0 & 64.0 & 70.0 \\
\textbf{2.0\%} & 28.0 & 45.0 & 60.0 & 74.0 \\
\bottomrule
\end{tabular} \\
\bottomrule
\end{tabular}%
% }
\caption{Accuracy results for different NeuroShield-ViT configurations on ViT-B/16-224 across different Subset sizes on Imagenet1k dataset. The table shows accuracy (\%) for various percentages of neurons neutralized (top-p) from each block in each layer, with different neutralizing coefficients.}
\label{tab:final_ablation_study}
\end{table*}

\subsection{CIFAR - 10 Accuracy}
On CIFAR-10 with ViT-B/16-224, NeuroShield-ViT significantly improved robustness, increasing accuracy from 23\% to 50\% for PGD-100 and 21\% to 49\% for PGD-20 attacks.

\subsection{Models}
\begin{itemize}
    \item ViT - \cite{dosovitskiy2020image}
    \item DeiT - \cite{touvron2021training}
\end{itemize}

\subsection{Datasets}
\begin{itemize}
    \item Imagenette - \cite{imagenette}
    \item Imagenet-1k - \cite{deng2009imagenet}
    \item  CIFAR - \cite{krizhevsky2010cifar}
\end{itemize}
\twocolumn
\bibliographystyle{IEEEtran}
\bibliography{main}

\end{document}